\documentclass[10pt,twocolumn,letterpaper]{article}

\usepackage[pagenumbers]{cvpr} 
\usepackage[dvipsnames]{xcolor}

\usepackage{multirow}

\usepackage[norelsize, linesnumbered, ruled, lined, boxed, commentsnumbered]{algorithm2e}
\definecolor{cvprblue}{rgb}{0.21,0.49,0.74}
\usepackage[pagebackref,breaklinks,colorlinks,citecolor=cvprblue]{hyperref}
\usepackage{diagbox}
\def\paperID{4144} 
\def\confName{CVPR}
\def\confYear{2024}

\title{Face Swap via Diffusion Model}

\author{Feifei Wang \\
{\tt\small {wangfeifei@mail.ustc.edu.cn}
}
}

\begin{document}
\maketitle

\section{Overview}

This technical report presents a diffusion model based framework for face swapping between two portrait images.
The basic framework consists of three components, \ie, IP-Adapter, ControlNet, and Stable Diffusion's inpainting pipeline, for face feature encoding, multi-conditional generation, and face inpainting respectively. 
Besides, I introduce facial guidance optimization and CodeFormer based blending to further improve the generation quality.

Specifically, we engage a recent light-weighted customization method (\ie, DreamBooth-LoRA), to guarantee the identity consistency by 
1) using a rare identifier ``sks'' to represent the source identity, and 
2) injecting the image features of source portrait into each cross-attention layer like the text features. 
Then I resort to the strong inpainting ability of Stable Diffusion, and utilize canny image and face detection annotation of the target portrait as the conditions, to guide ContorlNet's generation and align source portrait with the target portrait.
To further correct face alignment, we add the facial guidance loss to optimize the text embedding during the sample generation.


I conduct the experiments on CelebA-HQ to show the face alignment and identity fidelity quantitatively. 
Next, I give the details of each designs.

    
\section{Method}
\label{sec:formatting}
\subsection{Customization}
A lot of diffusion model based customization methods have been proposed recently, such as Textual Inversion \cite{gal2022textual}, DreamBooth \cite{ruiz2023dreambooth}, Custom Diffusion \cite{kumari2023multi} and their variants combined with LoRA \cite{hu2021lora}. 
Since the input is just one single portrait, I choose the light-weighted DreamBooth-LoRA, which only finetunes the low-ranked parameters added on the cross-attention layers in UNet. 
To avoid over-fitting, I set the training steps as 200. 
The prompt used during the training is ``a photo of sks person". 
For the prior knowledge during DreamBooth generation, I use 200 images generated by the pre-trained diffusion model with the prompt ``a photo of person".
\begin{figure}[h]
\includegraphics[width=1\linewidth]{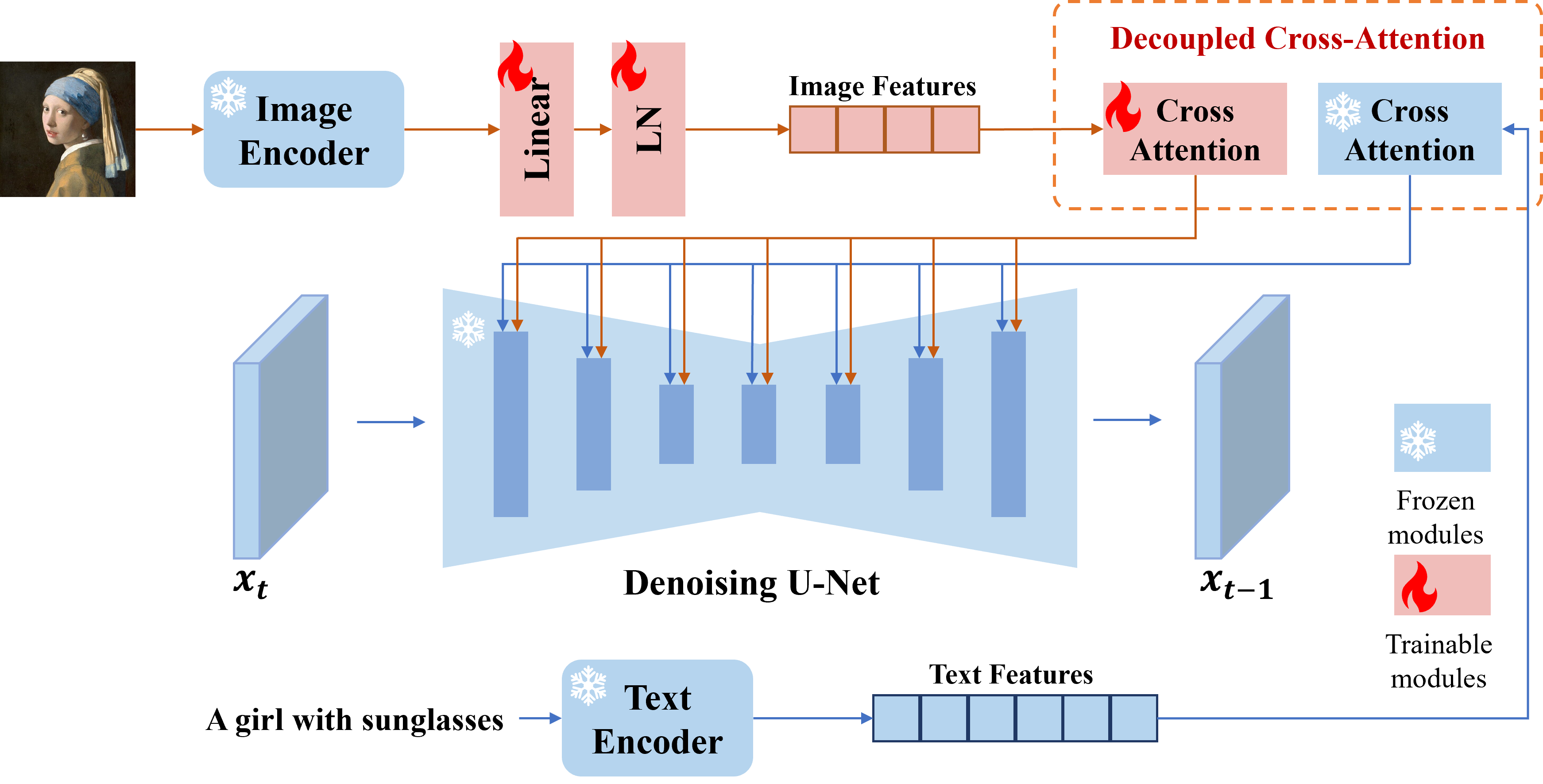}
\caption{The pipeline of IP-Adapter\cite{ye2023ip-adapter}}
\label{fig:ip-adapter}
\end{figure}

\subsection{Face Image Encoding}
Usually, an image contains more information than the corresponding text, thus DiffFace \cite{kim2022diffface} uses the embedding encoded by the ArcFace\cite{deng2019arcface} to guide the diffusion generation. 
However, the representations of the diffusion model are different from the face detection model. 
The pre-trained diffusion models are more familiar with text-image pairs, thus I adopt IP-Adapter\cite{ye2023ip-adapter} as an additional image encoder which uses the CLIP\cite{radford2021learning} vision model to preserve more identity information when generating the images. 
The pipeline of IP-Adapter is illustrated in Figure~\ref{fig:ip-adapter}.

\begin{figure}[h]
\begin{minipage}{0.312\linewidth}
    \centering
    \includegraphics[width=1\linewidth]{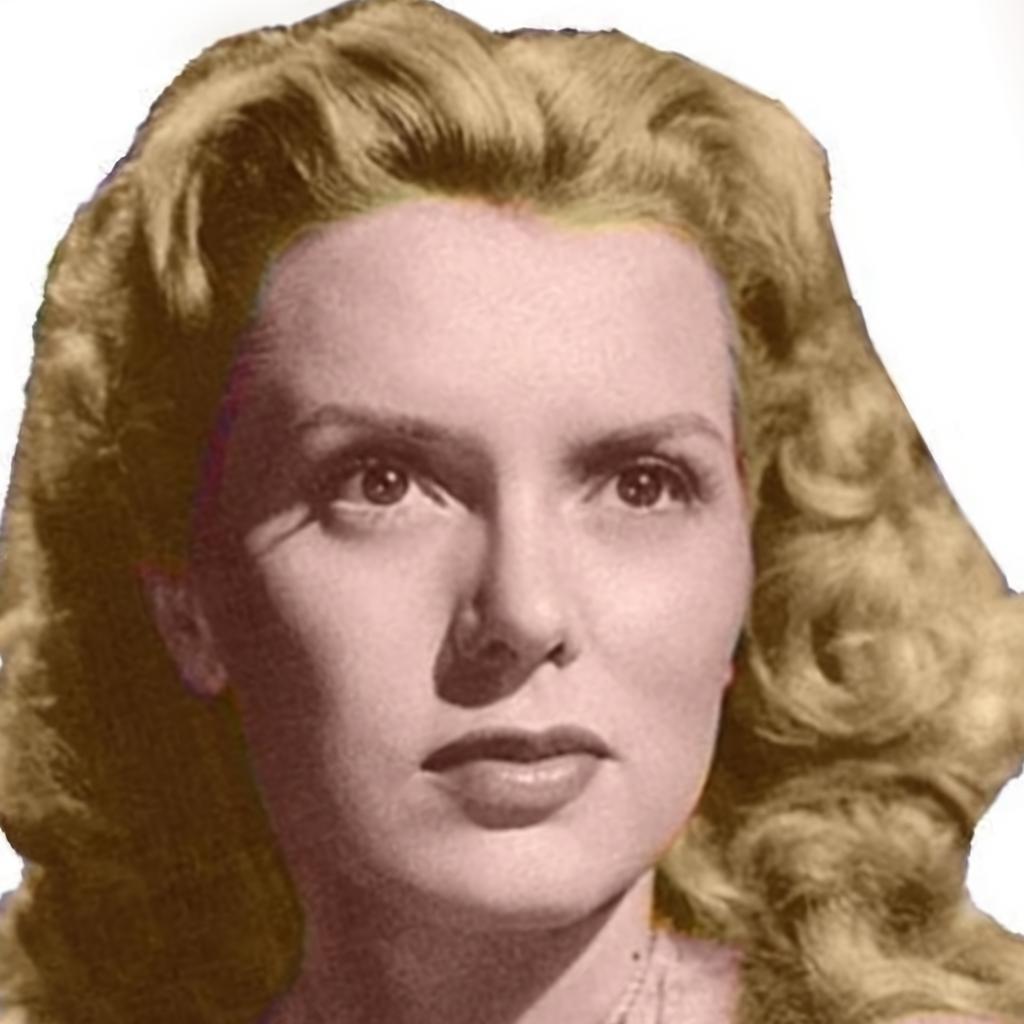}
    \centerline{\scriptsize \textbf {target image}}
\end{minipage}
\hfill
\begin{minipage}{0.312\linewidth}
    \centering
    \includegraphics[width=1\linewidth]{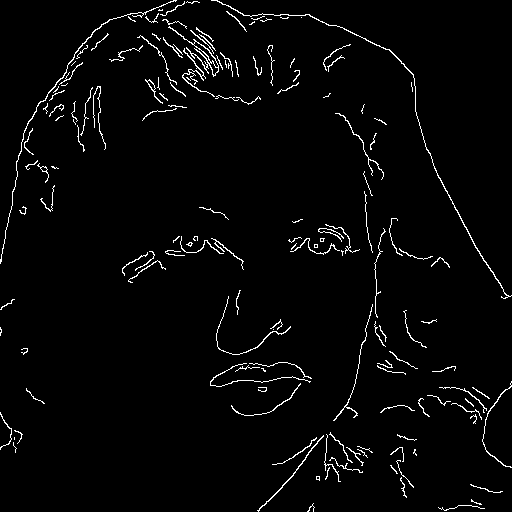}
    \centerline{\scriptsize \textbf{canny image}}
\end{minipage}
\hfill
\begin{minipage}{0.312\linewidth}
    \centering
    \includegraphics[width=1\linewidth]{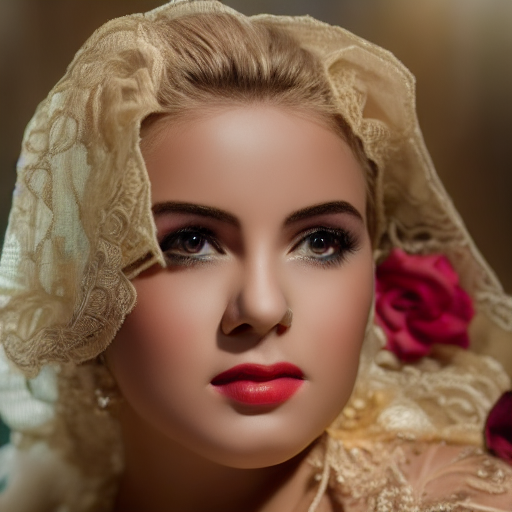}
    \centerline{\scriptsize  \textbf{result image}}
\end{minipage}
\caption{Samples generated by ControlNet\cite{zhang2023adding} v1.1 canny}
\label{fig:figure2}
\end{figure}

\begin{figure}[h]
\begin{minipage}{0.312\linewidth}
    \centering
    \includegraphics[width=1\linewidth]{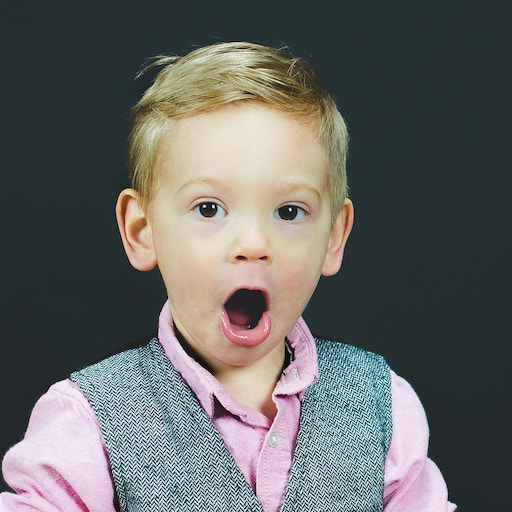}
    \centerline{\scriptsize \textbf {target image}}
\end{minipage}
\hfill
\begin{minipage}{0.312\linewidth}
    \centering
    \includegraphics[width=1\linewidth]{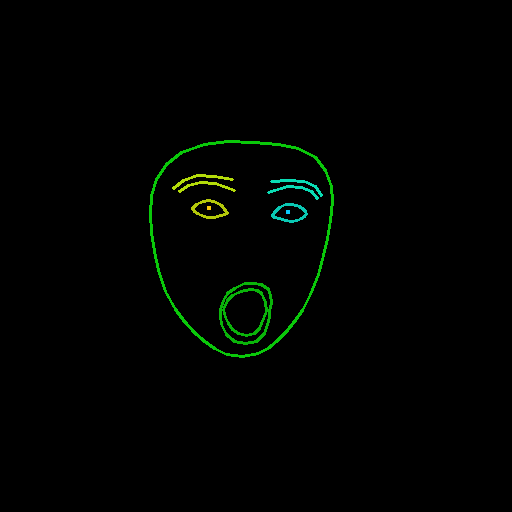}
    \centerline{\scriptsize \textbf{annotation image}}
\end{minipage}
\hfill
\begin{minipage}{0.312\linewidth}
    \centering
    \includegraphics[width=1\linewidth]{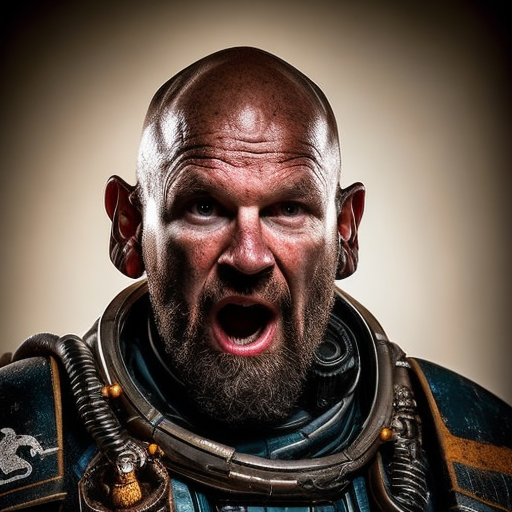}
    \centerline{\scriptsize  \textbf{result image}}
\end{minipage}
\caption{Samples generated by ControlNetMediaPipeFace\cite{face}}
\label{fig:figure3}
\end{figure}

\subsection{Multi-Conditional Generation}
ControlNet\cite{zhang2023adding} acts as a new paradigm for condition image generation. It can support multiple condition inputs. 
Crucible AI\cite{face} trains a ControlNet with human facial expressions which can allow users to control the human face generation. 

However, only taking facial detection annotation (Figure~\ref{fig:figure3}) as a condition often results in the face shape misalignment between the source and target, making the swapped results unnatural. 
Thus, I adopt canny input (Figure~\ref{fig:figure2}) as another condition to help improve the alignment. 
Also, Only using canny images as conditions may lead to performance degradation and the resulting faces cannot match the target pose. 
Therefore, it is better to combine canny images with annotation images as the overall generation guidance.

\subsection{Face Mask and Inpainting}
Here our objective is to seamlessly perform face swapping. 
The principle is that only the facial area should be replaced and the remains must be reserved. 
I borrow the strong inpainting ability of the open-sourced Stable Diffusion models and obtain the face mask with advanced face detection models. 
In general, I combine the Stable Diffusion inpainting pipeline with the ControlNet and IP-Adapter, thus the output can align the target portrait images with the identity of the source portrait.

\subsection{Facial Guidance Optimization}
During sampling, the expression of generated faces can be heavily affected by the LoRA-added weights, which may lead to the over-fitting issue of facial details.  

Since text embedding is crucial in generating desired facial details, I find that adding appropriate words in the text prompt can modify facial expressions. 
Inspired by Null-text inversion and Imagic, I optimize the text embedding to improve similarity and correct the facial expression during inference. 

As defined in DDIM \cite{ddim}, the sample $x_{t-1}$ is transited from sample $x_{t}$ and pred original samples $\hat{x}_{0}$ in Eq ~\ref{pred}:

\begin{equation}\label{pred}
\begin{split}
x_{t-1}&=\sqrt{\alpha_{t-1}}\hat{x}_{0}+\sqrt{1-\alpha_{t-1}-\sigma_t^2} \cdot \epsilon_\theta^{(t)}\left(x_t\right)\\
\hat{x}_{0} &= \frac{x_t-\sqrt{1-\alpha_t} \epsilon_\theta^{(t)}\left(x_t\right)}{\sqrt{\alpha_t}}
\end{split}
\end{equation}

where $x_{t}$ is current sample and $x_{t-1}$ is the previous sample, $\hat{x}_{0}$ is the predicted original samples and $\epsilon_\theta^{(t)}$ is the corresponding model output. Since the facial models are trained for clean images, I extract the predicted $\hat{x}_{0}$. 

Similar to DiffFace\cite{kim2022diffface}, I input it to face parsing model $F_{P}$ and output a facial parsing map with essential labels like skin, eyes, and eyebrows. During sampling, I compute the distance of the face semantic feature between predicted $\hat{x}_{0}$ and the target portrait. 

\begin{equation}\label{db}     
\begin{split}
\mathcal{L}_{p} = \|F_{P}(x_{targ}) - F_{P}(\hat{x_{0}})\|_2^2 
\end{split}
\end{equation}


\begin{figure}[t]
\begin{minipage}{0.24\linewidth}
    \centering
    \includegraphics[width=1\linewidth]{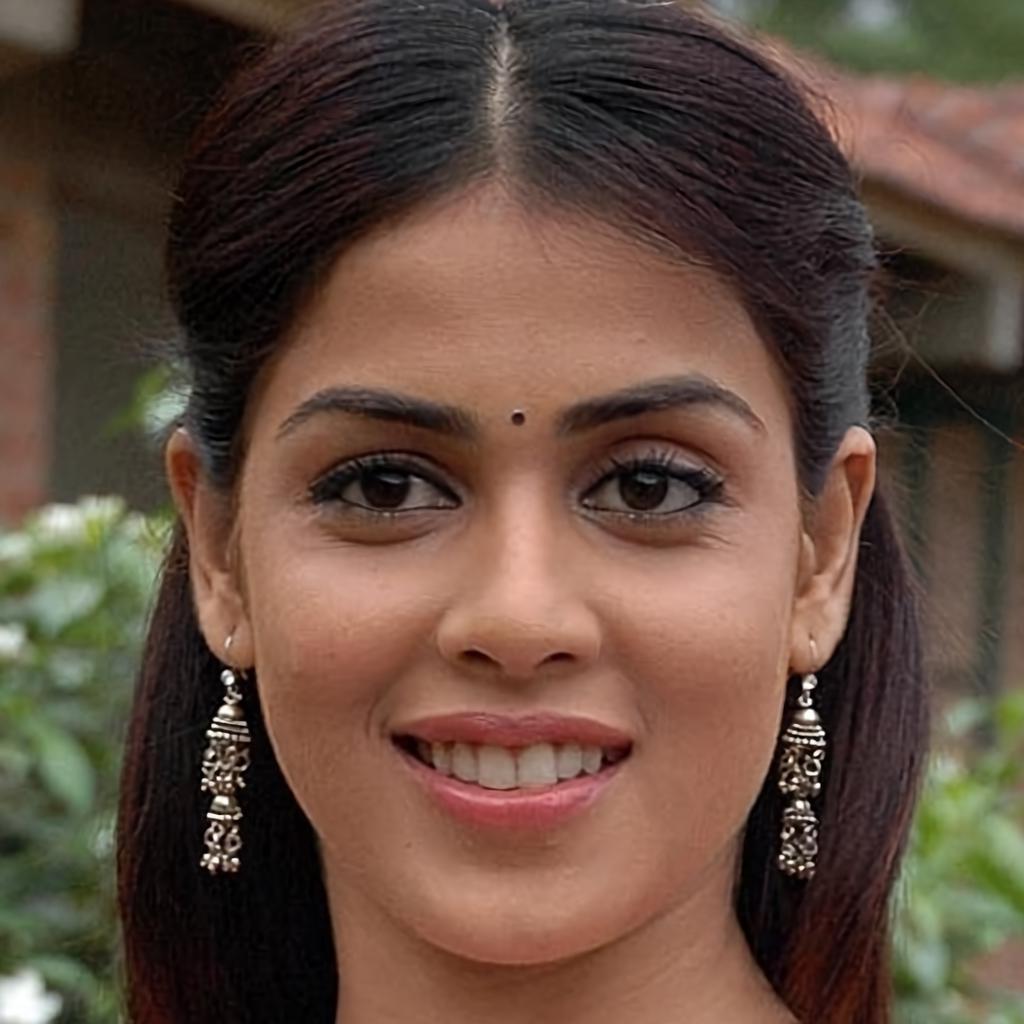}
    \centerline{\scriptsize \textbf {source portrait}}
\end{minipage}
\begin{minipage}{0.24\linewidth}
    \centering
    \includegraphics[width=1\linewidth]{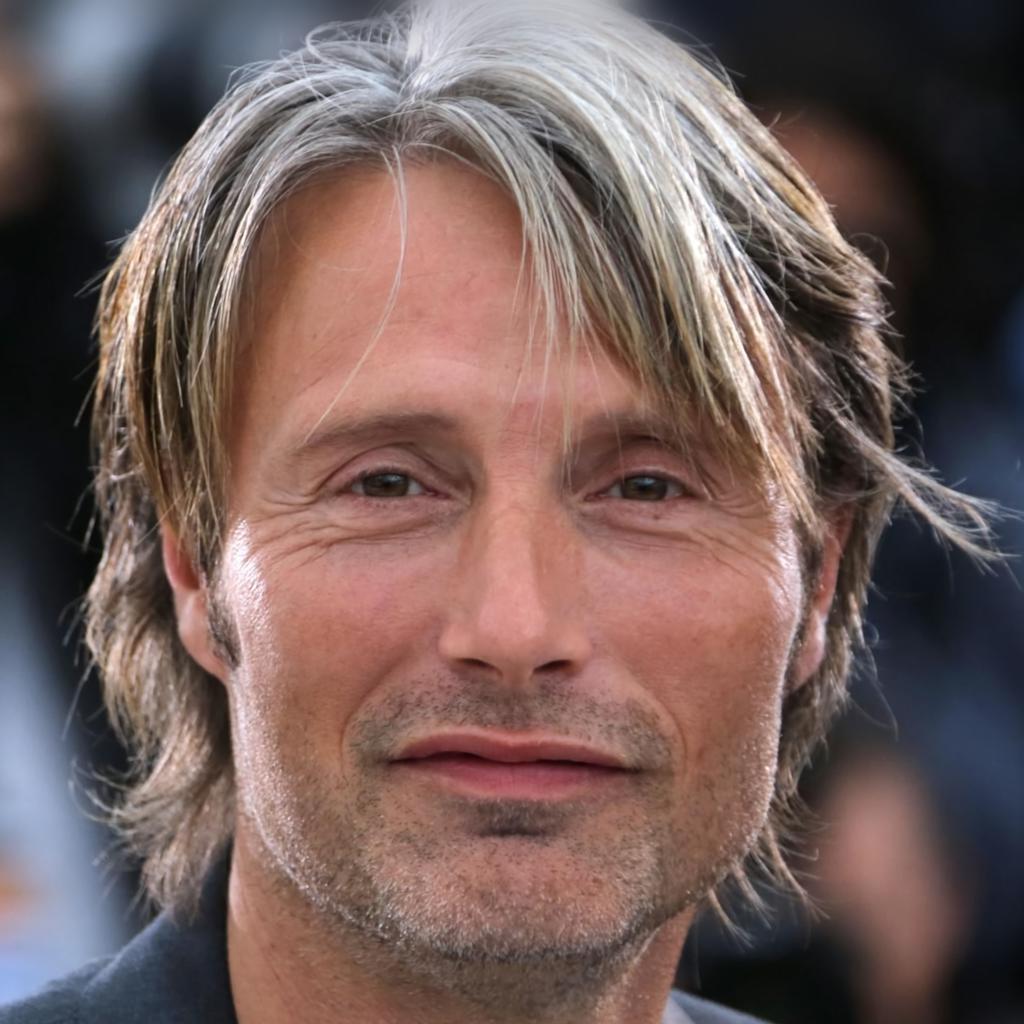}
    \centerline{\scriptsize \textbf {target portrait}}
\end{minipage}
\begin{minipage}{0.24\linewidth}
    \centering
    \includegraphics[width=1\linewidth]{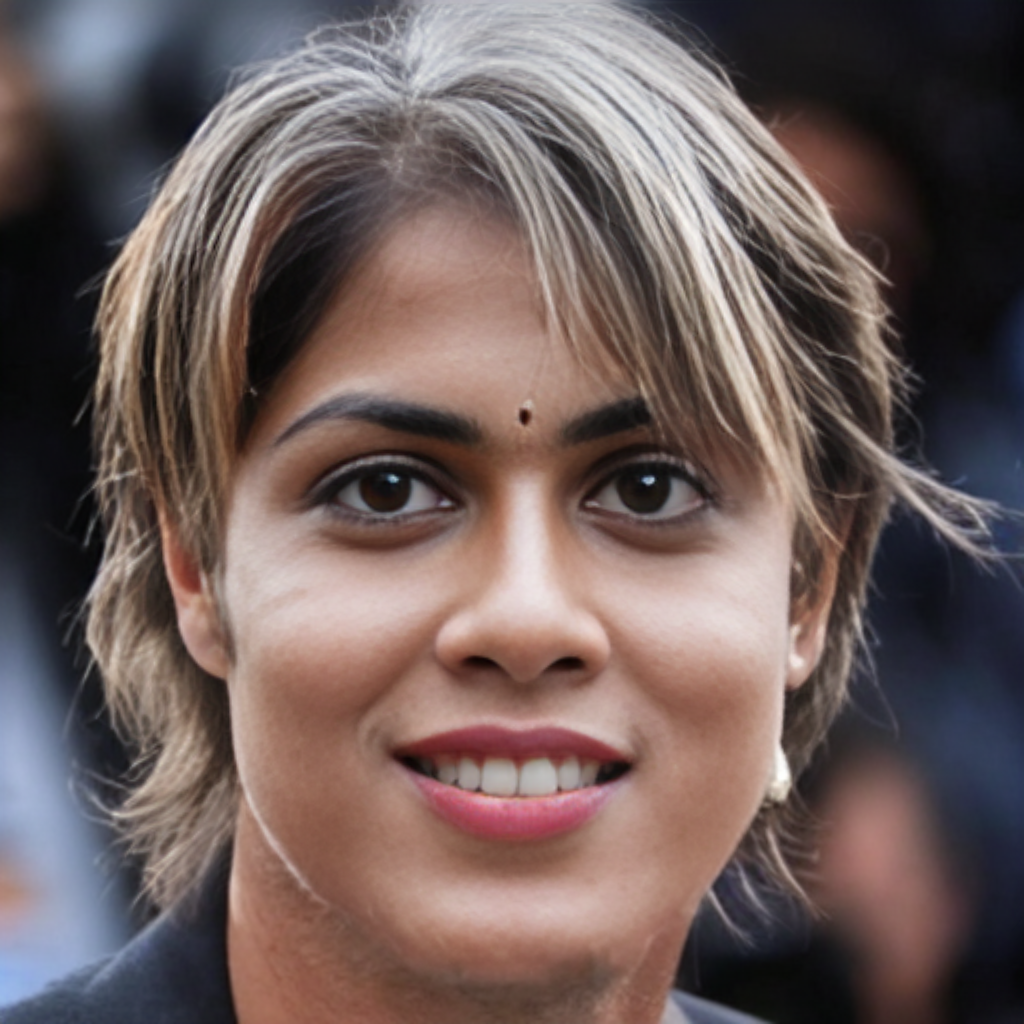}
    \centerline{\scriptsize \textbf {w/o loss}}
\end{minipage}
\begin{minipage}{0.24\linewidth}
    \centering
    \includegraphics[width=1\linewidth]{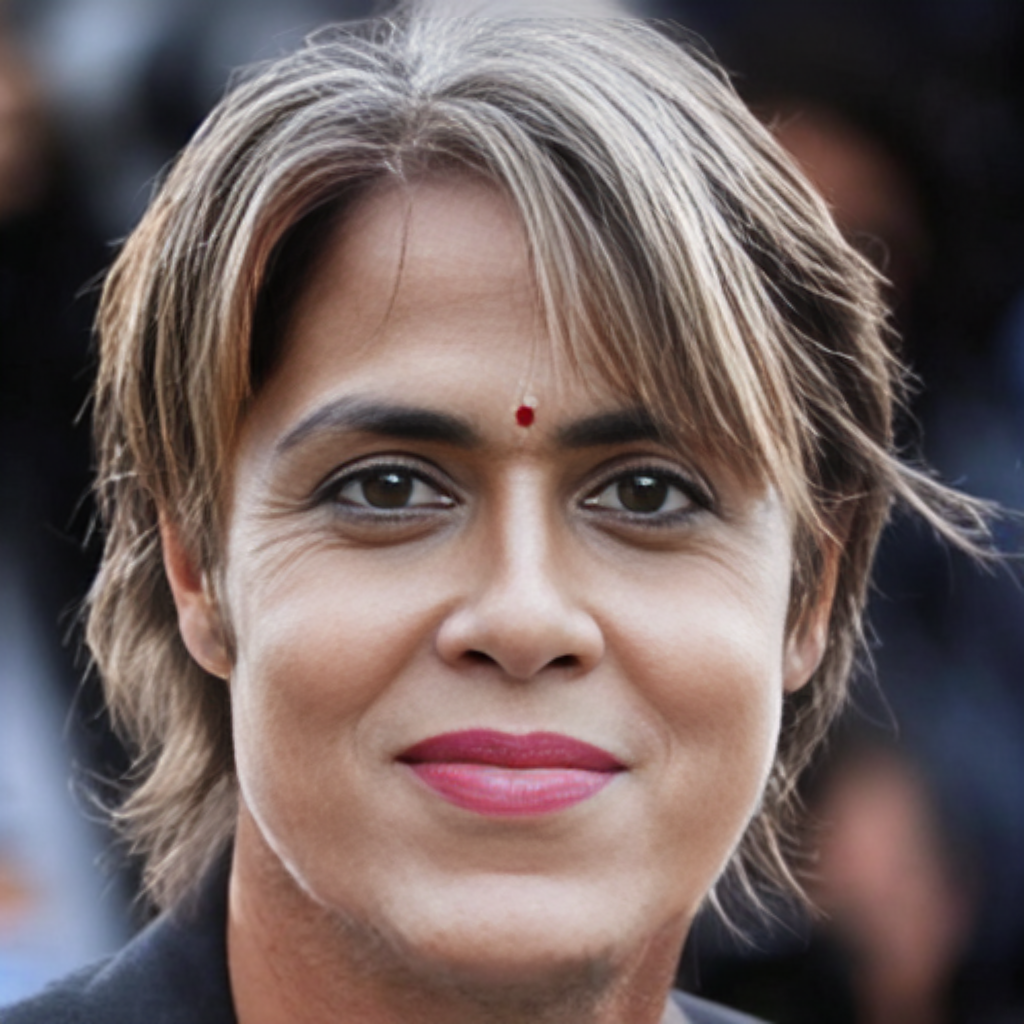}
    \centerline{\scriptsize \textbf {w loss}}
\end{minipage}

\caption{Samples generated by facial guidance loss}
\label{fig:opt}
\end{figure}

Fig~\ref{fig:opt} shows that optimizing the conditional text embedding can improve the face alignment.

\subsection{Face Restoration} Since the inpainting sometimes cannot seemly merge the generated face and the background, I need some tools to make the results more natural. CodeFormer\cite{zhou2022codeformer} is an expert face enhancement model that is widely used in stable-diffusion webui\cite{sd-wb} and the results are usually excellent.

\section{Experiments}
I randomly select 100 image pairs from the Celeb-HQ to conduct the evaluation. The source and target are both single portraits and the results are compared with DiffFace baseline.

\subsection{Quantitative Evaluation}
There are two aspects to measure the generated face quality. The identity is measured by ID-embedder. Since DiffFace\cite{kim2022diffface} training uses the ArcFace\cite{deng2019arcface}, I take the CosFace\cite{wang2018cosface} to encode image embedding and compute source and target cosine similarity. To evaluate shape, expression, and pose, I follow \cite{kim2022smooth} to use \cite{4d} to get the expression, pose, and shape params of the 3D Face model\cite{3d} and compute the L1 distances between source and target params. Compared with DiffFace, the results in Tab ~\ref{tab:compare} indicate that our face-swapping results have a closer resemblance to the target facial expressions, pose, and shape, but there is a slight decrease in facial similarity. In the generation process, there's a balance between preserving identity and achieving face-swapping. Therefore, we need to introduce additional constraints to further enhance identity consistency.

\begin{table}[t]
\centering
\begin{tabular}{l|c|ccc}
\midrule
{Method}& Cos$\uparrow$& Expr$\downarrow$ & Pose $\downarrow$& Shape $\downarrow$ \\
\midrule
DiffFace\cite{kim2022diffface} & 0.4841 &0.0357 &0.0178  &0.0642 
\\
Our &0.4020 &0.0263 &0.0153 &0.0512
\\
\bottomrule
\end{tabular}
\caption{Comparison with DiffFace on the subset of CelebA-HQ dataset. I evaluate the performance of ID cosine similarity, expression, pose, and shape, where higher cosine similarity is better and lower expression, pose and shape distances are better.}
\label{tab:compare}
\end{table}
\subsection{Qualitative Evaluation}
\begin{figure}[h]
\begin{minipage}{0.24\linewidth}
    \centering
    \includegraphics[width=1\linewidth]{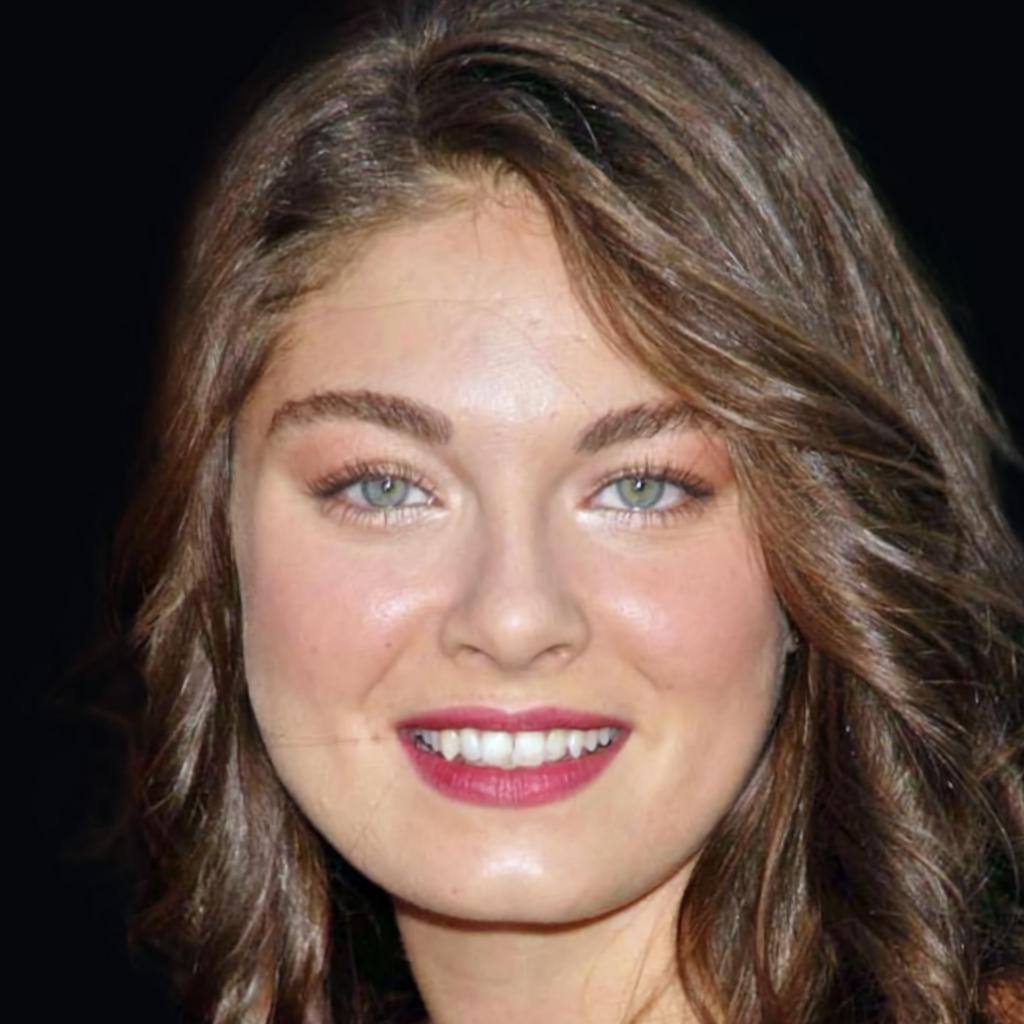}
\end{minipage}
\begin{minipage}{0.24\linewidth}
    \centering
    \includegraphics[width=1\linewidth]{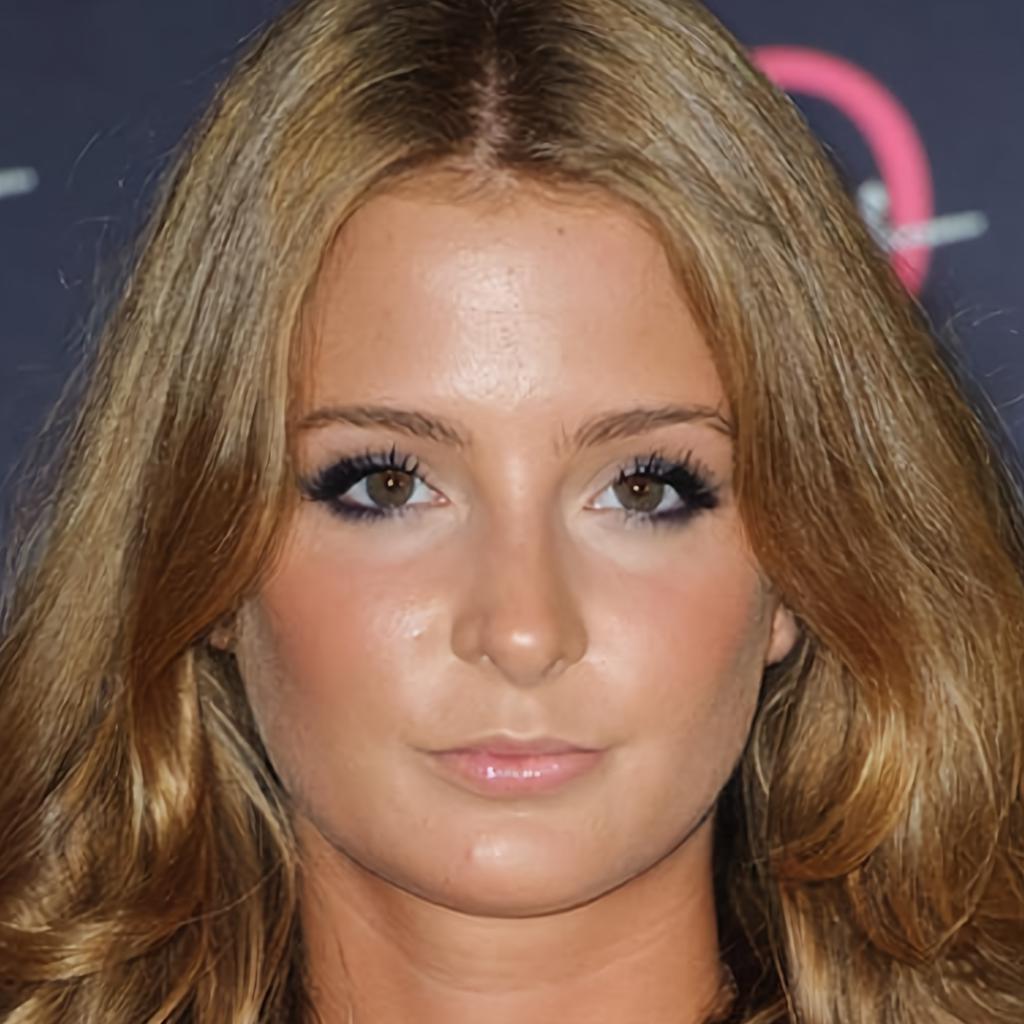}
\end{minipage}
\begin{minipage}{0.24\linewidth}
    \centering
    \includegraphics[width=1\linewidth]{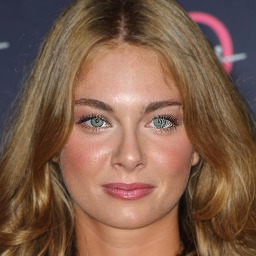}
\end{minipage}
\begin{minipage}{0.24\linewidth}
    \centering
    \includegraphics[width=1\linewidth]{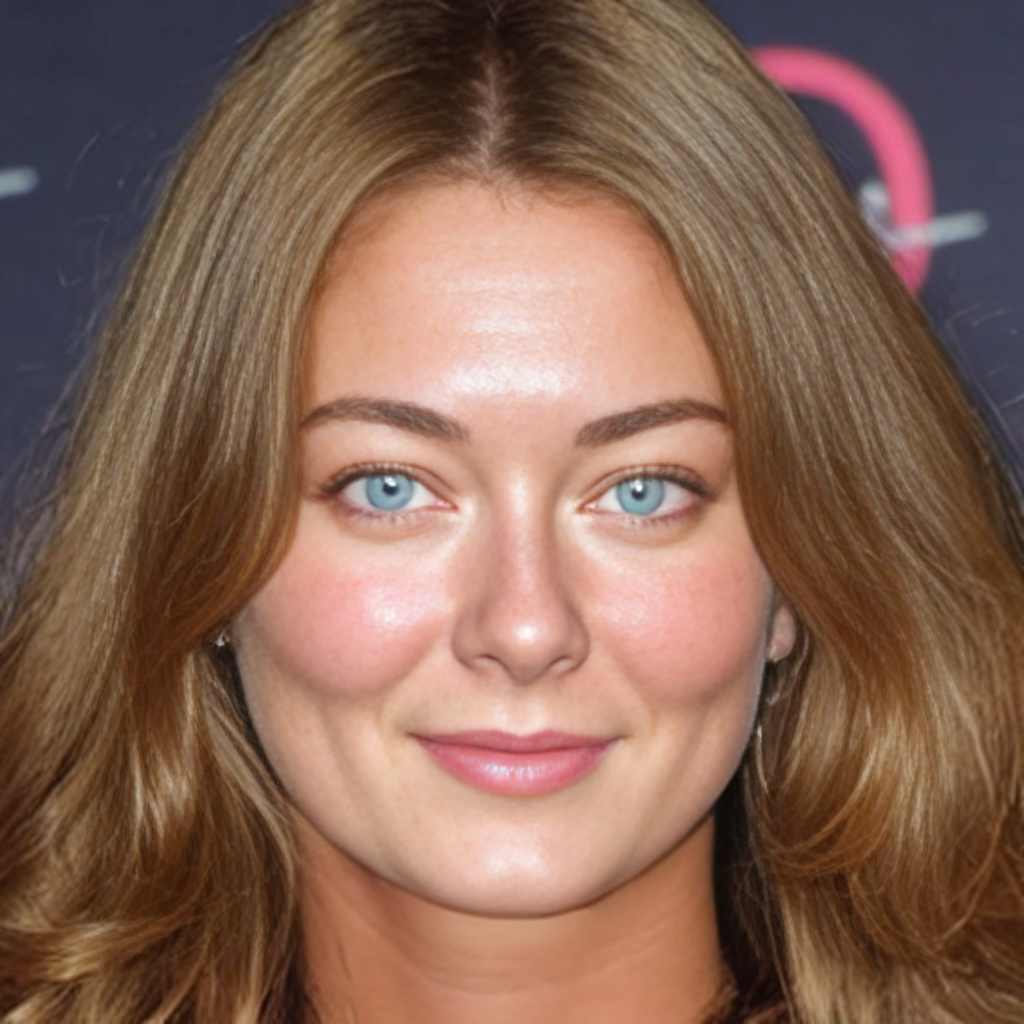}
\end{minipage}
\begin{minipage}{0.24\linewidth}
    \centering
    \includegraphics[width=1\linewidth]{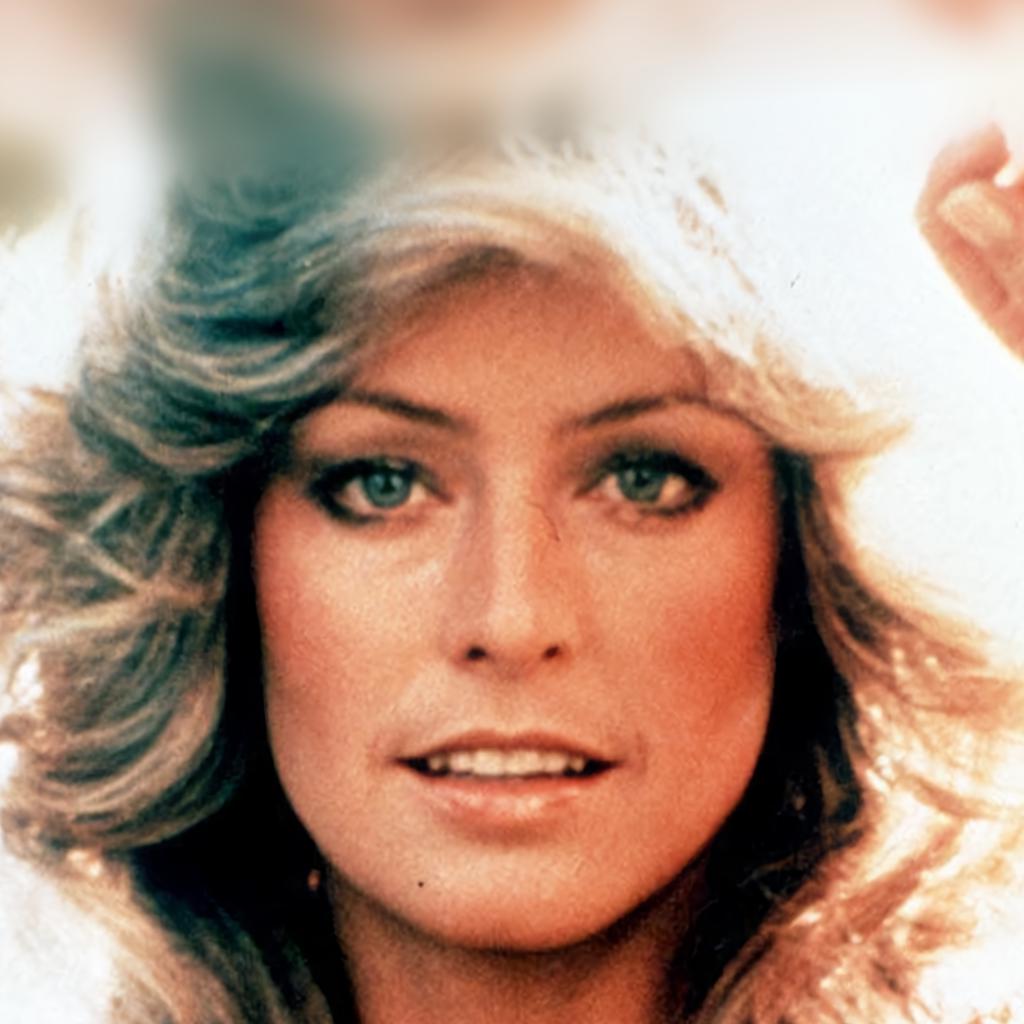}
\end{minipage}
\begin{minipage}{0.24\linewidth}
    \centering
    \includegraphics[width=1\linewidth]{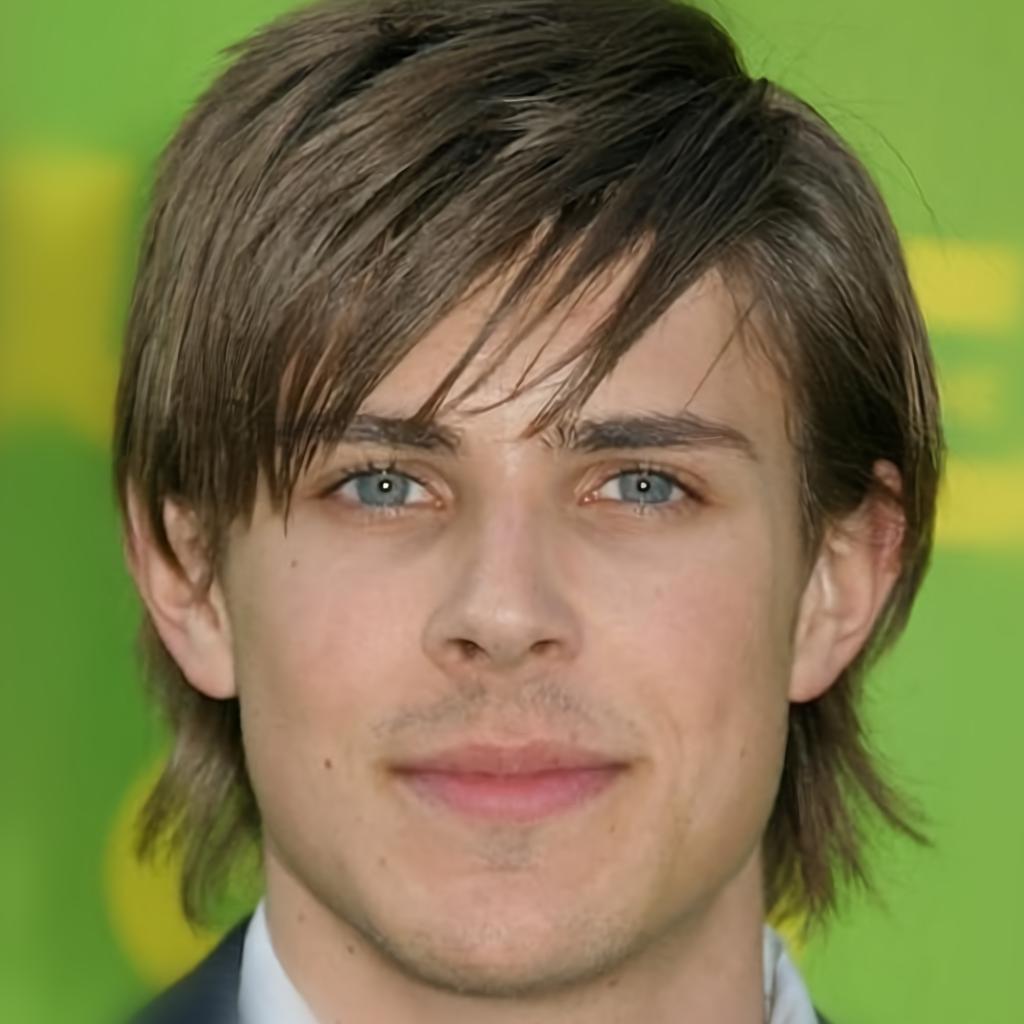}
\end{minipage}
\begin{minipage}{0.24\linewidth}
    \centering
    \includegraphics[width=1\linewidth]{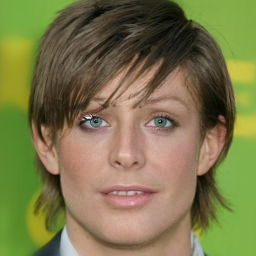}
\end{minipage}
\begin{minipage}{0.24\linewidth}
    \centering
    \includegraphics[width=1\linewidth]{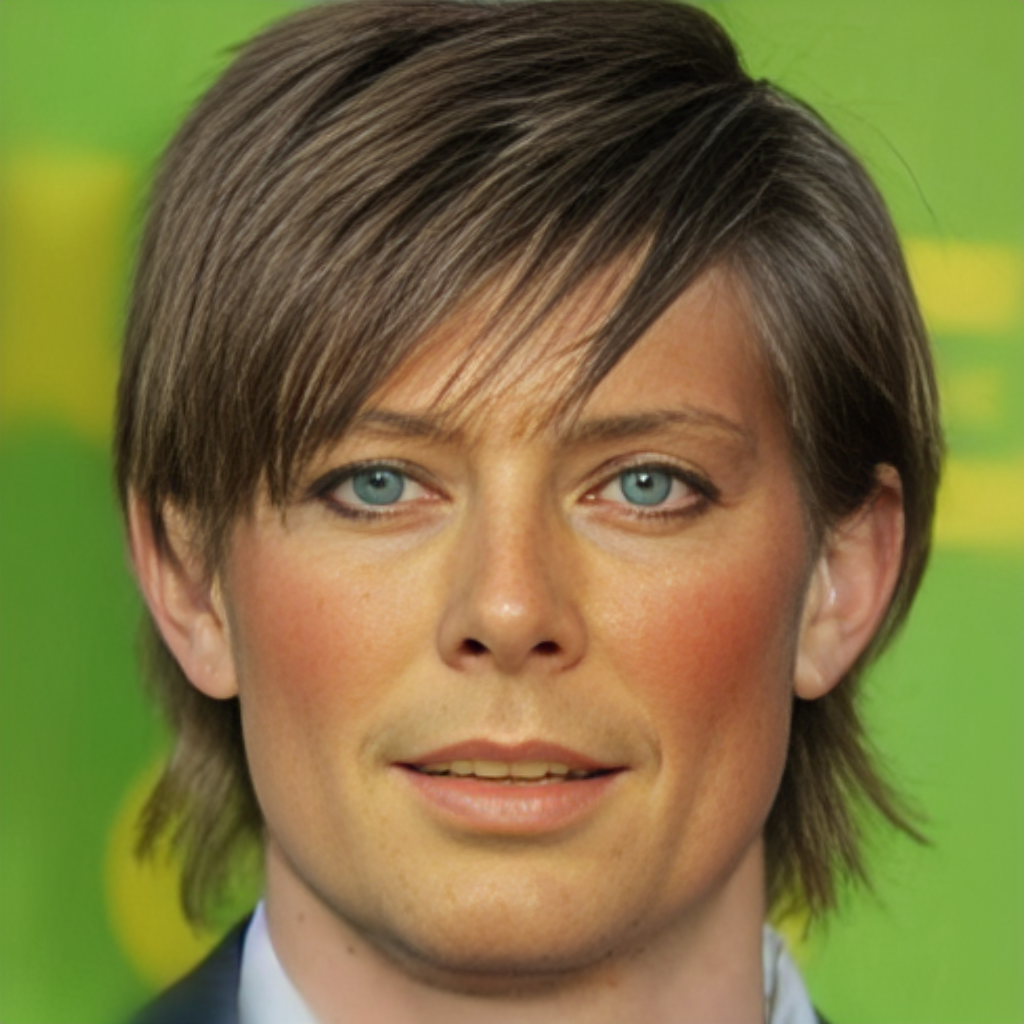}
\end{minipage}
\begin{minipage}{0.24\linewidth}
    \centering
    \includegraphics[width=1\linewidth]{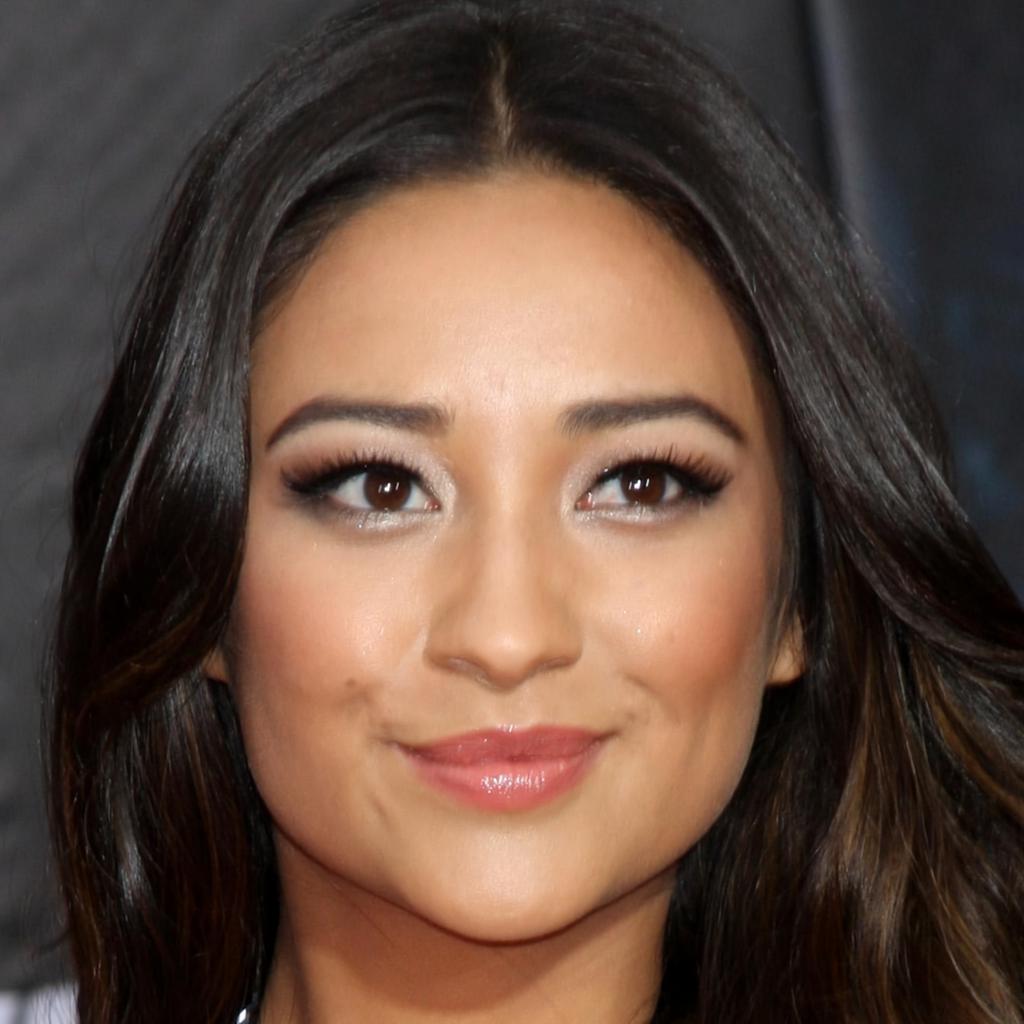}
\end{minipage}
\begin{minipage}{0.24\linewidth}
    \centering
    \includegraphics[width=1\linewidth]{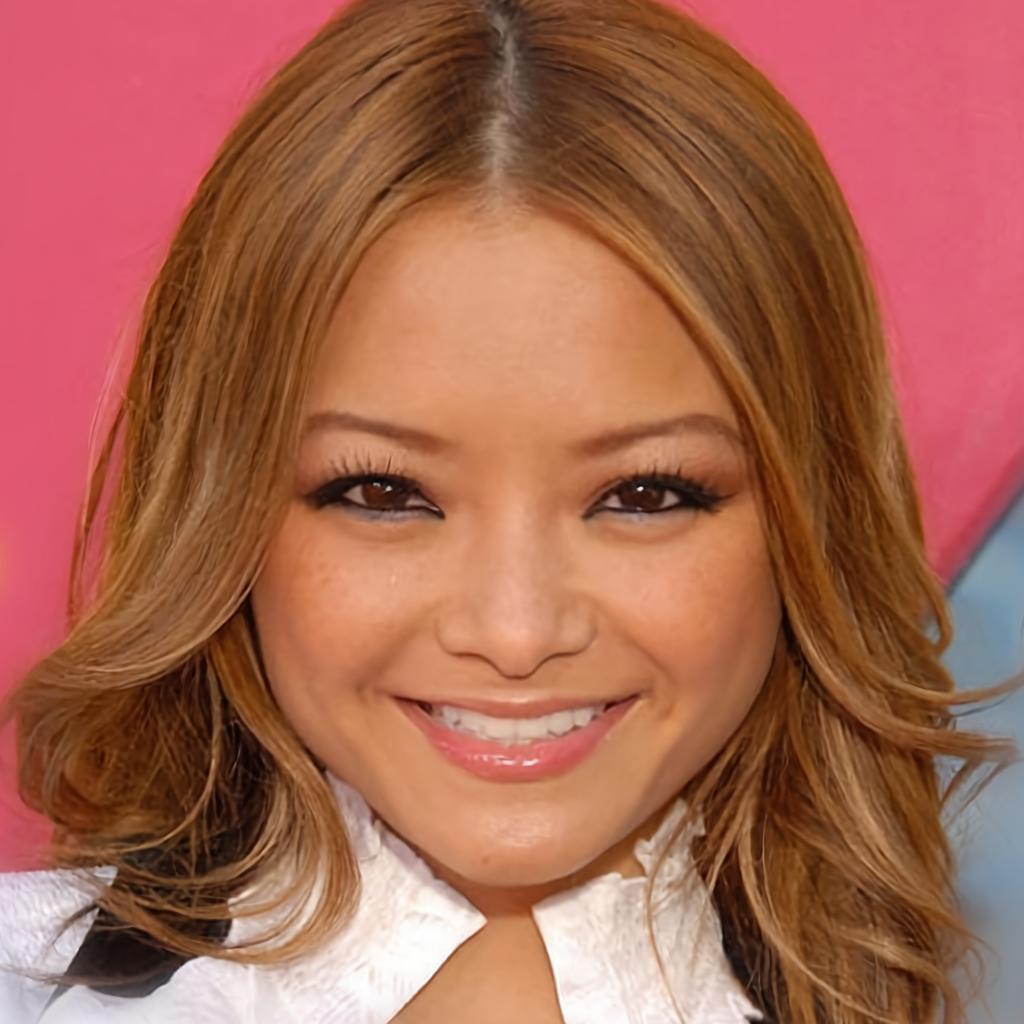}
\end{minipage}
\begin{minipage}{0.24\linewidth}
    \centering
    \includegraphics[width=1\linewidth]{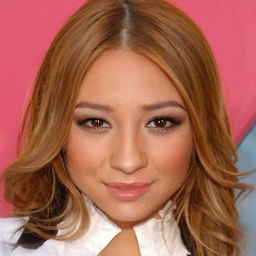}
\end{minipage}
\begin{minipage}{0.24\linewidth}
    \centering
    \includegraphics[width=1\linewidth]{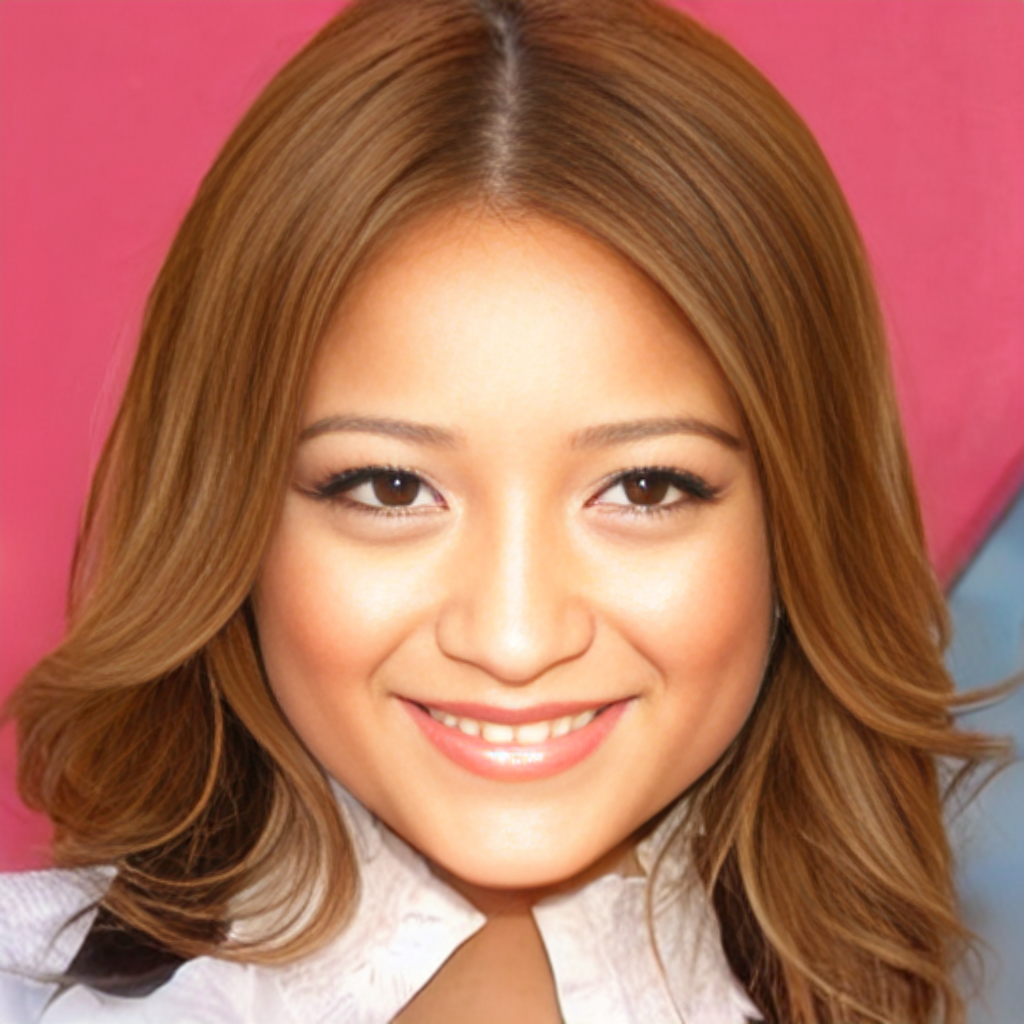}
\end{minipage}
\begin{minipage}{0.24\linewidth}
    \centering
    \includegraphics[width=1\linewidth]{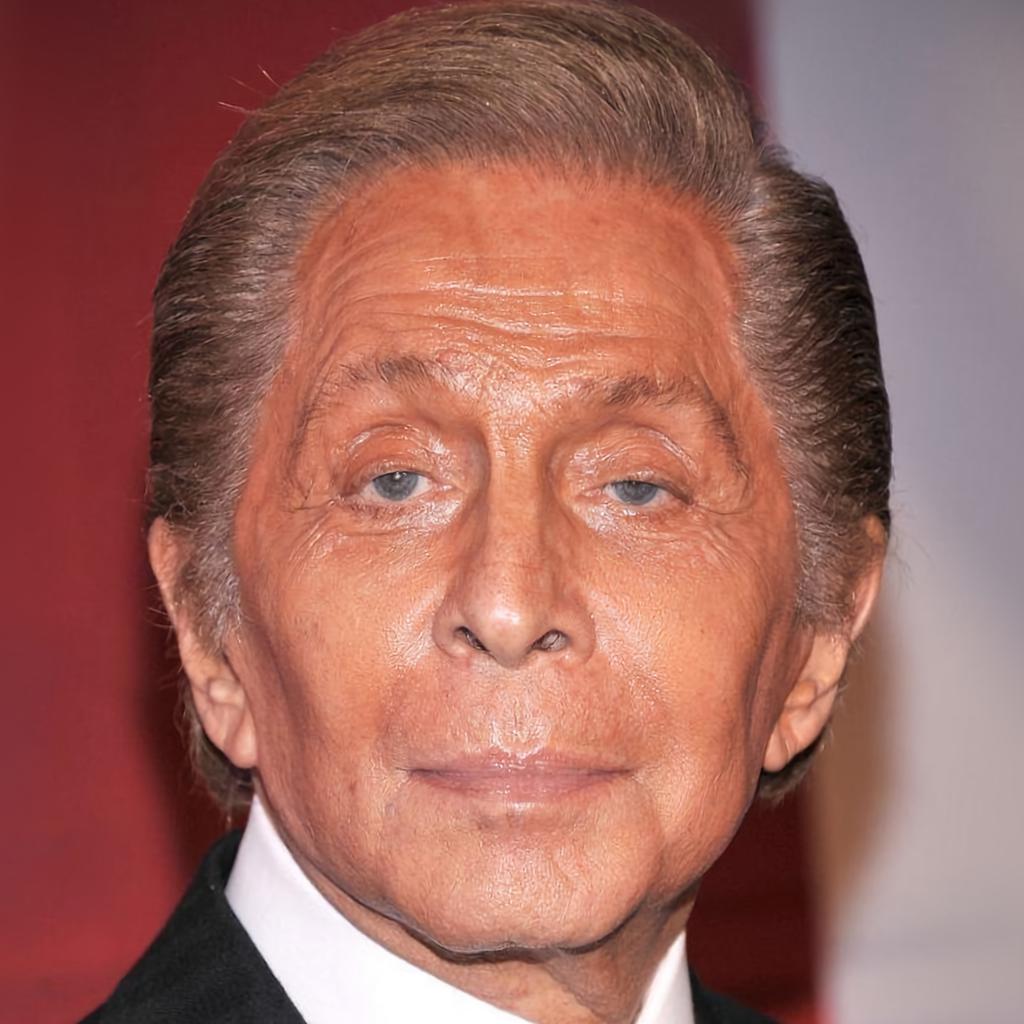}
\end{minipage}
\begin{minipage}{0.24\linewidth}
    \centering
    \includegraphics[width=1\linewidth]{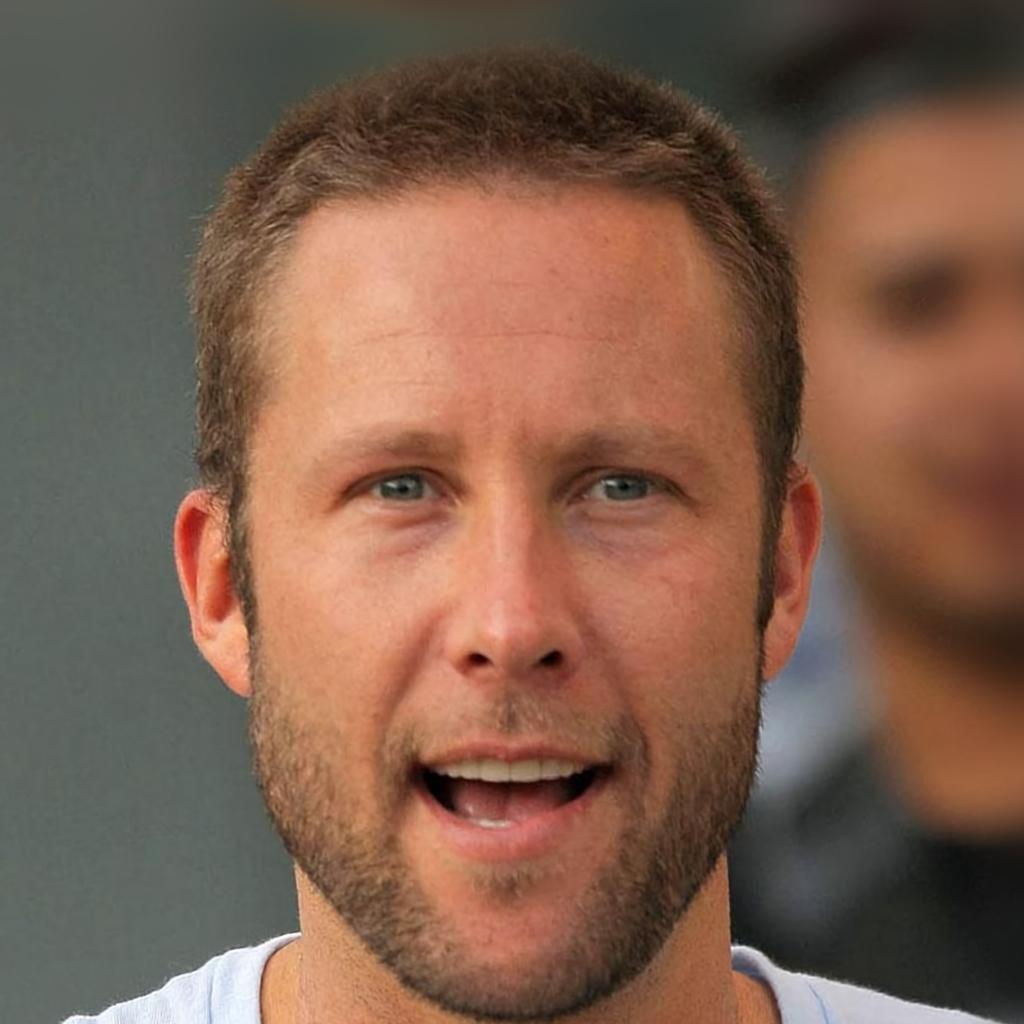}
\end{minipage}
\begin{minipage}{0.24\linewidth}
    \centering
    \includegraphics[width=1\linewidth]{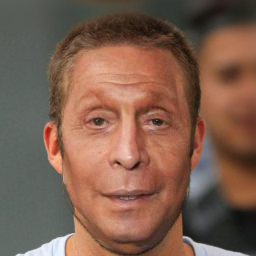}
\end{minipage}
\begin{minipage}{0.24\linewidth}
    \centering
    \includegraphics[width=1\linewidth]{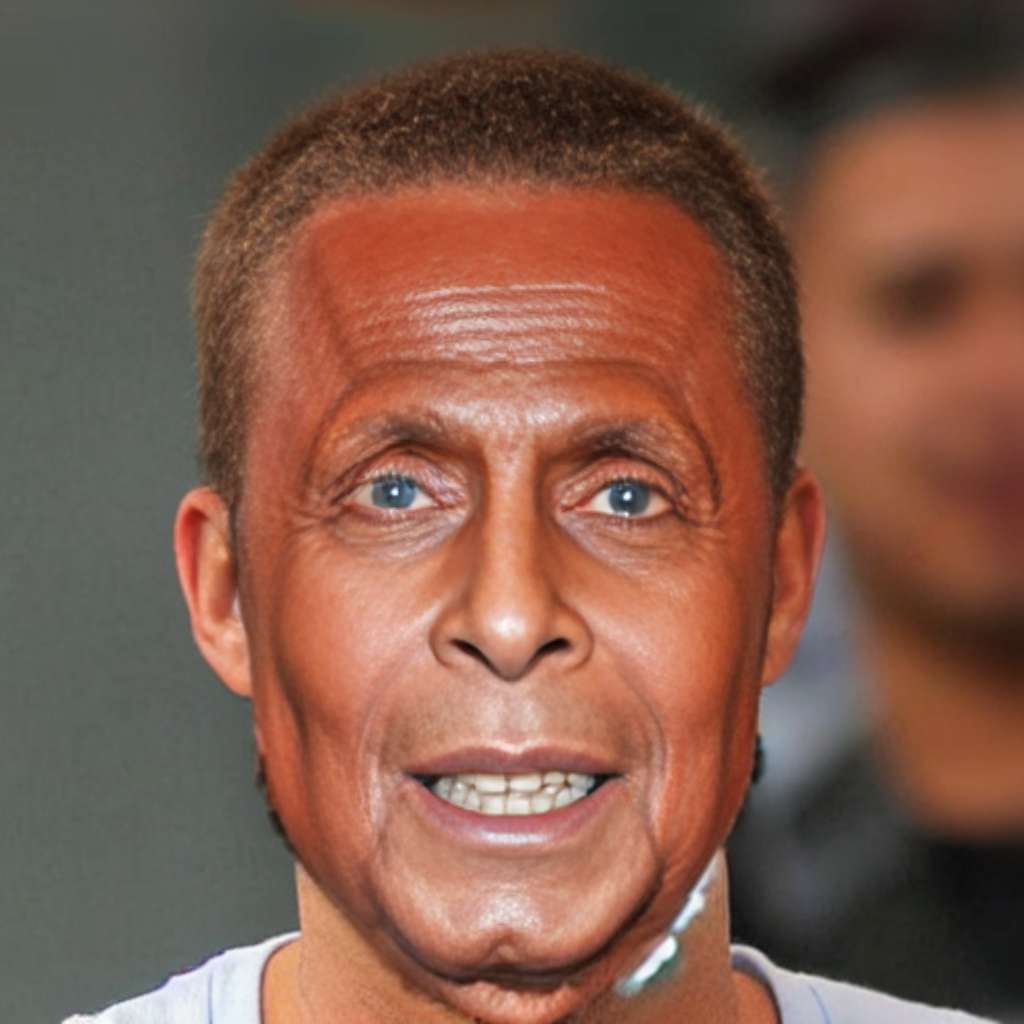}
\end{minipage}
\begin{minipage}{0.24\linewidth}
    \centering
    \includegraphics[width=1\linewidth]{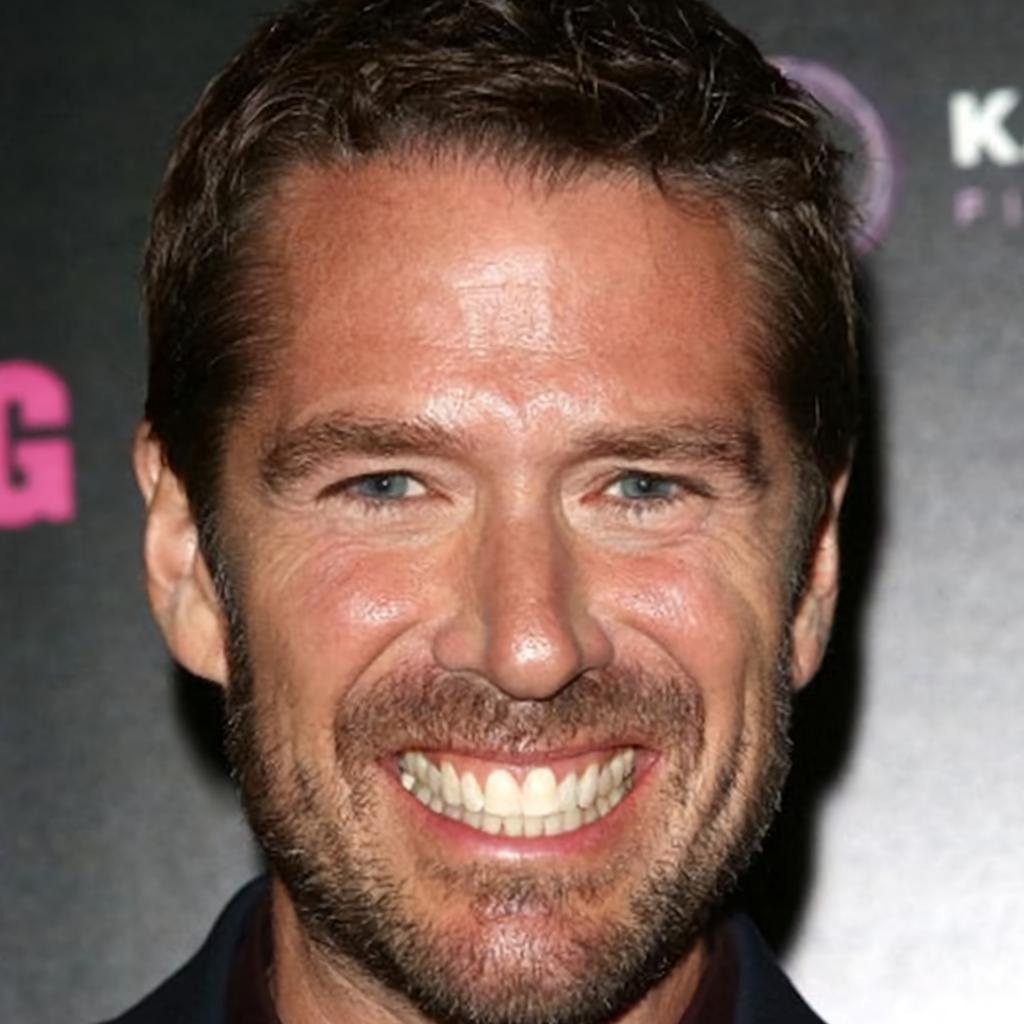}
\end{minipage}
\begin{minipage}{0.24\linewidth}
    \centering
    \includegraphics[width=1\linewidth]{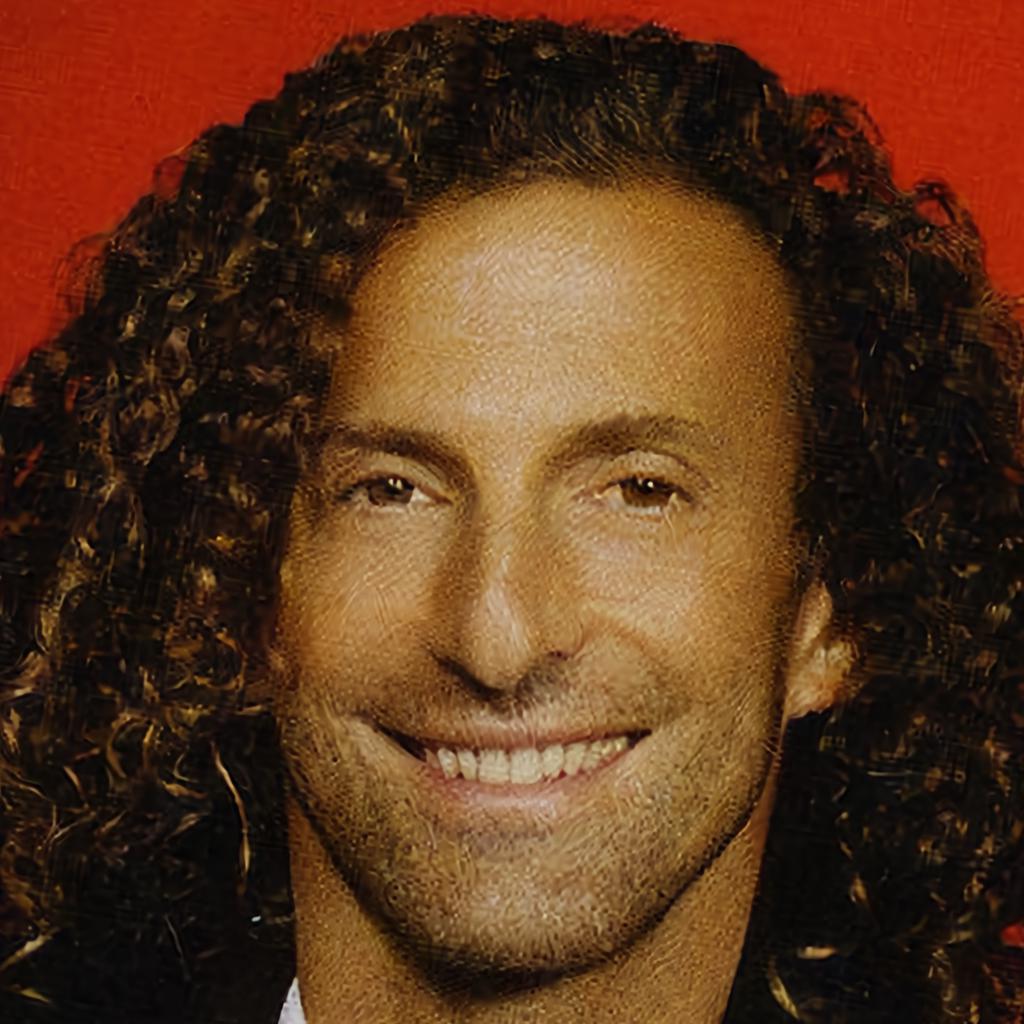}
\end{minipage}
\begin{minipage}{0.24\linewidth}
    \centering
    \includegraphics[width=1\linewidth]{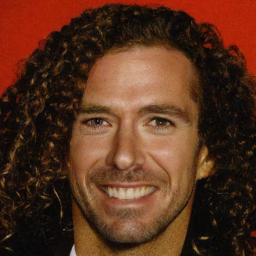}
\end{minipage}
\begin{minipage}{0.24\linewidth}
    \centering
    \includegraphics[width=1\linewidth]{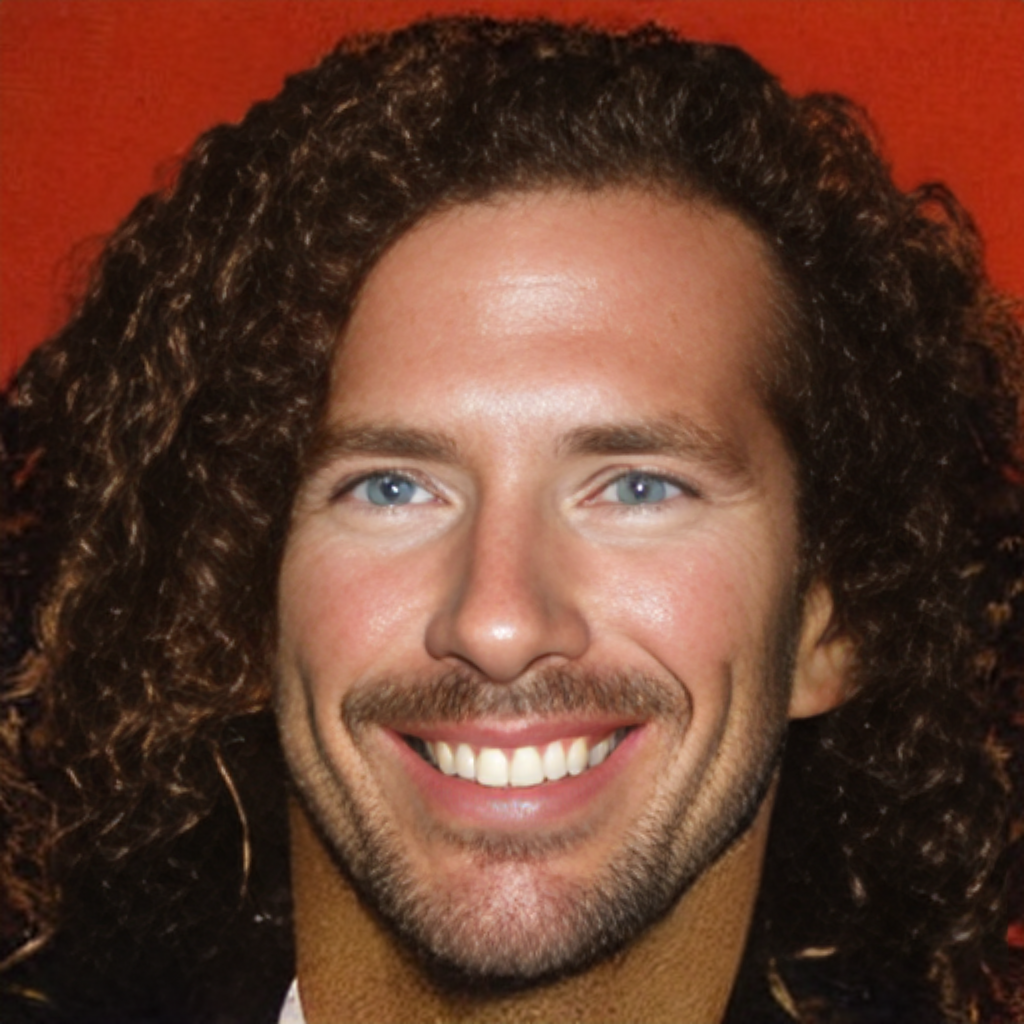}
\end{minipage}
\begin{minipage}{0.24\linewidth}
    \centering
    \includegraphics[width=1\linewidth]{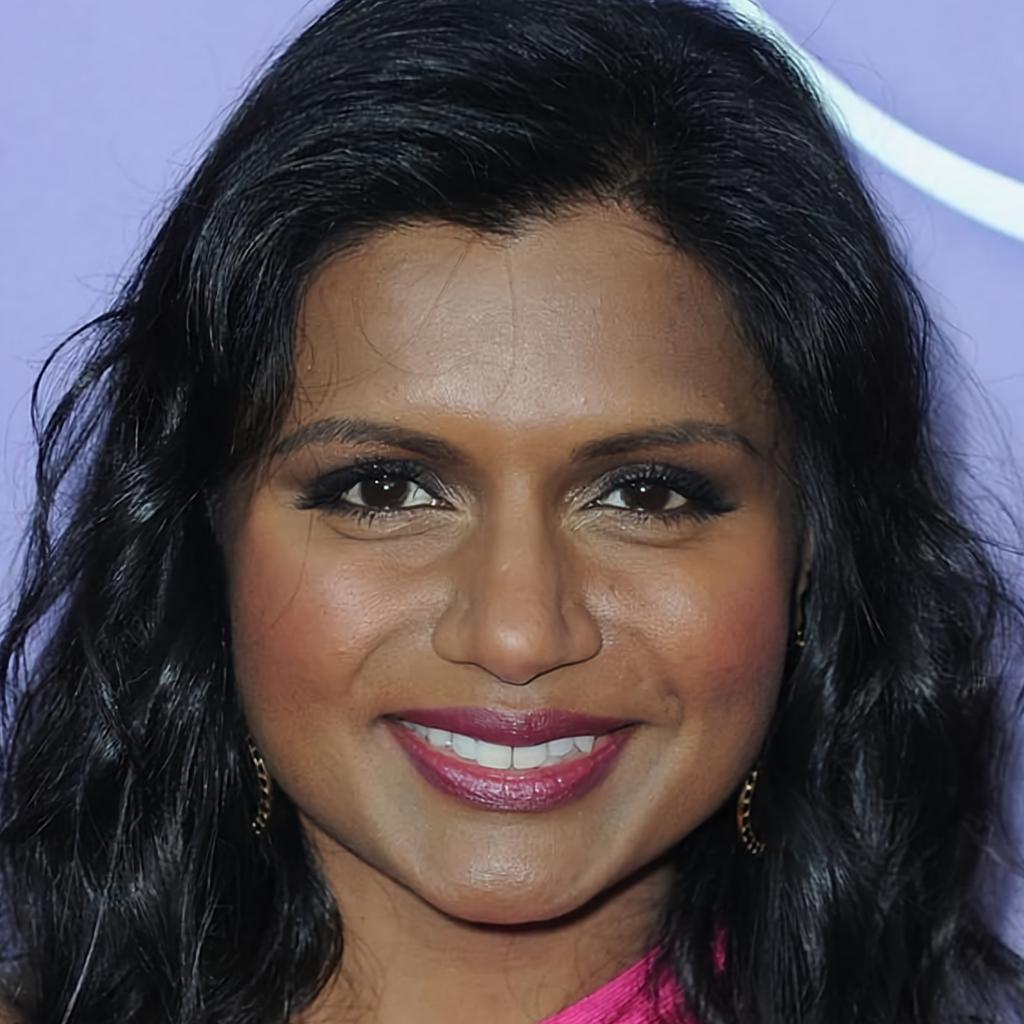}
     \centerline{\scriptsize \textbf {source portrait}}
\end{minipage}
\begin{minipage}{0.24\linewidth}
    \centering
    \includegraphics[width=1\linewidth]{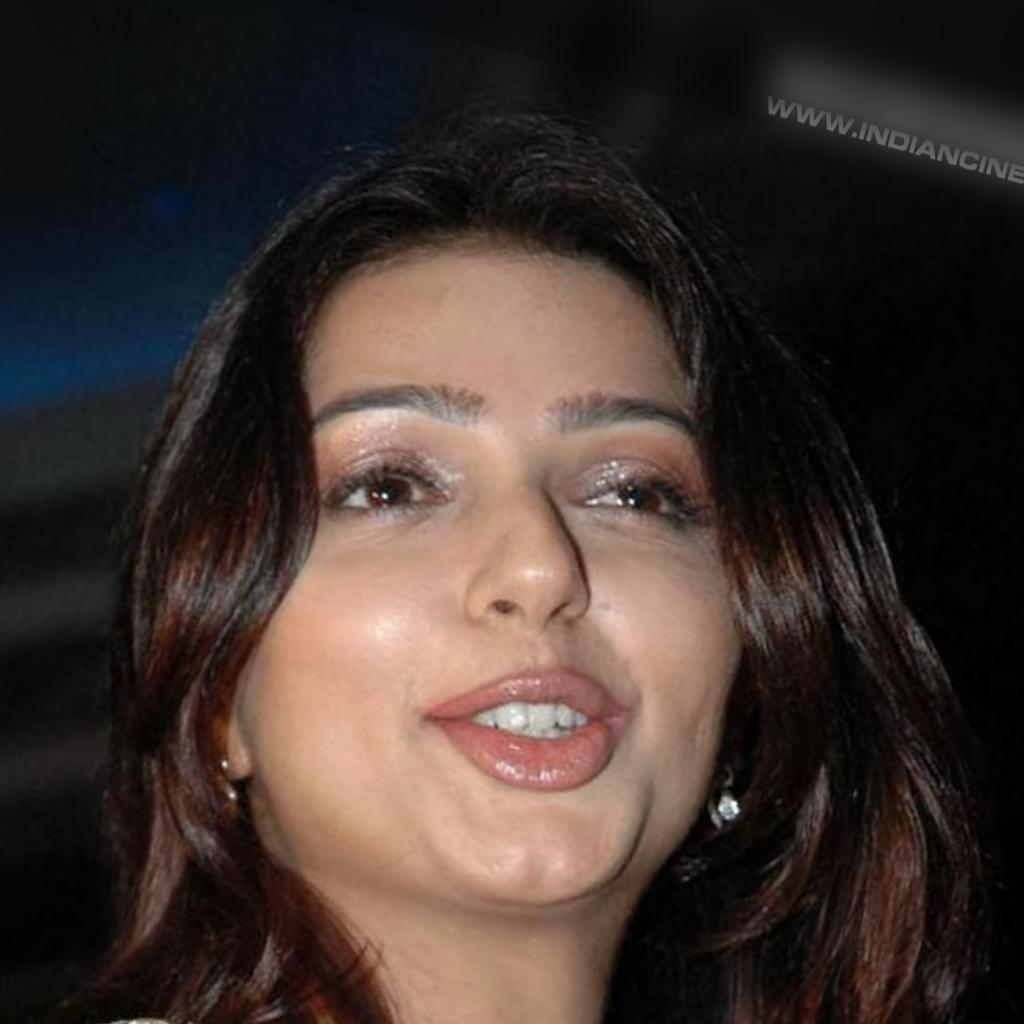}
    \centerline{\scriptsize \textbf {target portrait}}
\end{minipage}
\begin{minipage}{0.24\linewidth}
    \centering
    \includegraphics[width=1\linewidth]{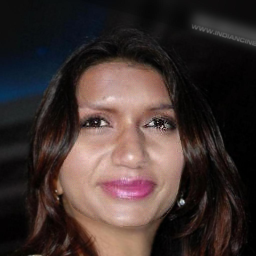}
    \centerline{\scriptsize \textbf {DiffFace}}
\end{minipage}
\begin{minipage}{0.24\linewidth}
    \centering
    \includegraphics[width=1\linewidth]{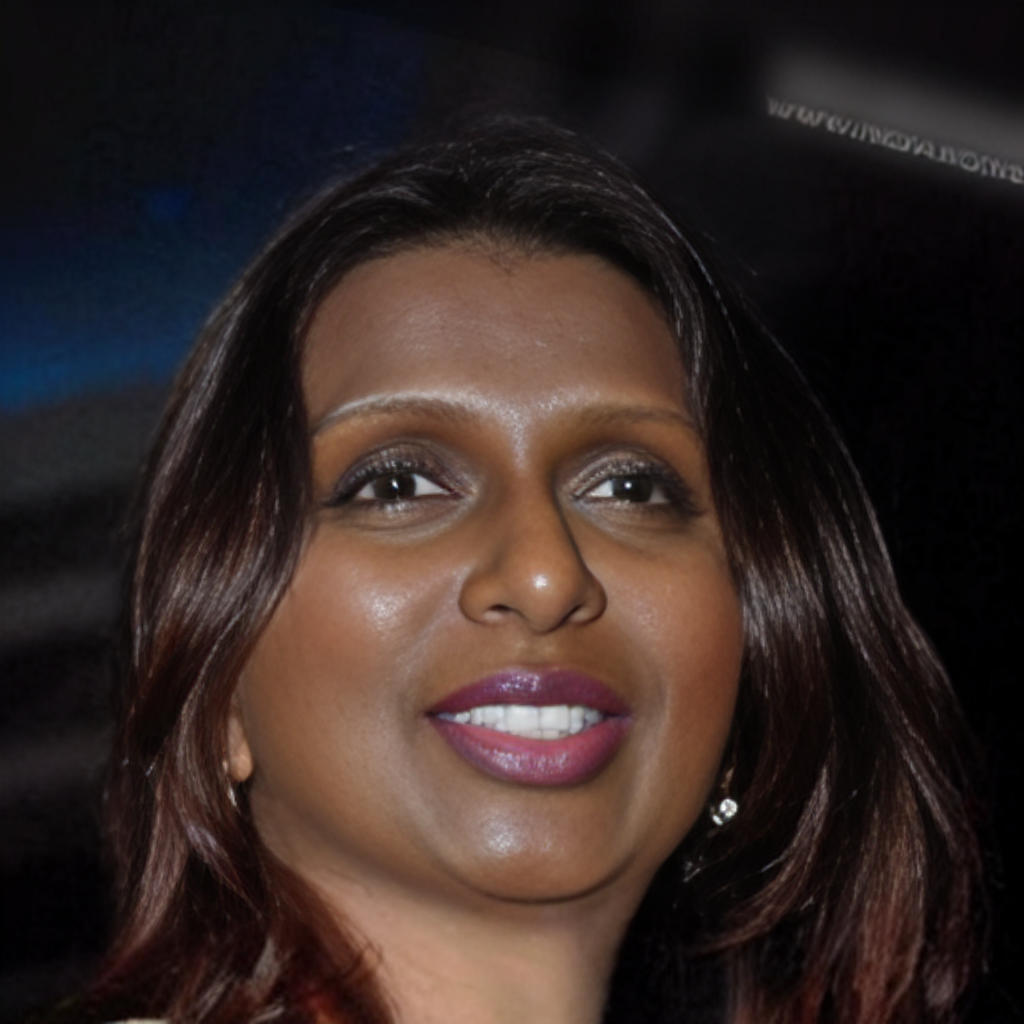}
    \centerline{\scriptsize \textbf {our}}
\end{minipage}
\caption{Quantitative results compared with DiffFace}
\label{fig:compare}
\end{figure}

As shown in Fig ~\ref{fig:compare}, our qualitative results are consistent with the quantitative evaluation.  ControlNet\cite{zhang2023adding} condition blocks and facial guidance optimization make better facial alignment with the target portrait. However, due to insufficient stable diffusion inpainting effects, the naturalness of the swapped faces is not so good. Therefore, further improvement is needed and quantitative metrics like face x-ray\cite{li2020face} can be employed to comprehensively assess the effectiveness of face-swapping.

{
    \small
    \bibliographystyle{ieeenat_fullname}
    \bibliography{main}

\begin{thebibliography}{18}
\providecommand{\natexlab}[1]{#1}
\providecommand{\url}[1]{\texttt{#1}}
\expandafter\ifx\csname urlstyle\endcsname\relax
  \providecommand{\doi}[1]{doi: #1}\else
  \providecommand{\doi}{doi: \begingroup \urlstyle{rm}\Url}\fi

\bibitem[AUTOMATIC1111()]{sd-wb}
AUTOMATIC1111.
\newblock stable-diffusion-webui.

\bibitem[CrucibleAI()]{face}
CrucibleAI.
\newblock Controlnetmediapipeface.

\bibitem[Deng et~al.(2019)Deng, Guo, Xue, and Zafeiriou]{deng2019arcface}
Jiankang Deng, Jia Guo, Niannan Xue, and Stefanos Zafeiriou.
\newblock Arcface: Additive angular margin loss for deep face recognition.
\newblock In \emph{Proceedings of the IEEE/CVF Conference on Computer Vision and Pattern Recognition}, pages 4690--4699, 2019.

\bibitem[Gal et~al.(2022)Gal, Alaluf, Atzmon, Patashnik, Bermano, Chechik, and Cohen-Or]{gal2022textual}
Rinon Gal, Yuval Alaluf, Yuval Atzmon, Or Patashnik, Amit~H. Bermano, Gal Chechik, and Daniel Cohen-Or.
\newblock An image is worth one word: Personalizing text-to-image generation using textual inversion, 2022.

\bibitem[Hu et~al.(2021)Hu, Shen, Wallis, Allen-Zhu, Li, Wang, Wang, and Chen]{hu2021lora}
Edward~J Hu, Yelong Shen, Phillip Wallis, Zeyuan Allen-Zhu, Yuanzhi Li, Shean Wang, Lu Wang, and Weizhu Chen.
\newblock Lora: Low-rank adaptation of large language models.
\newblock \emph{arXiv preprint arXiv:2106.09685}, 2021.

\bibitem[Kim et~al.(2022{\natexlab{a}})Kim, Lee, and Zhang]{kim2022smooth}
Jiseob Kim, Jihoon Lee, and Byoung-Tak Zhang.
\newblock Smooth-swap: a simple enhancement for face-swapping with smoothness.
\newblock In \emph{Proceedings of the IEEE/CVF Conference on Computer Vision and Pattern Recognition}, pages 10779--10788, 2022{\natexlab{a}}.

\bibitem[Kim et~al.(2022{\natexlab{b}})Kim, Kim, Cho, Seo, Nam, Lee, Kim, and Lee]{kim2022diffface}
K. Kim, Y. Kim, S. Cho, J. Seo, J. Nam, K. Lee, S. Kim, and K. Lee.
\newblock Diffface: Diffusion-based face swapping with facial guidance.
\newblock \emph{Arxiv}, 2022{\natexlab{b}}.

\bibitem[Kumari et~al.(2023)Kumari, Zhang, Zhang, Shechtman, and Zhu]{kumari2023multi}
Nupur Kumari, Bingliang Zhang, Richard Zhang, Eli Shechtman, and Jun-Yan Zhu.
\newblock Multi-concept customization of text-to-image diffusion.
\newblock In \emph{Proceedings of the IEEE/CVF Conference on Computer Vision and Pattern Recognition}, pages 1931--1941, 2023.

\bibitem[Li et~al.(2020)Li, Bao, Zhang, Yang, Chen, Wen, and Guo]{li2020face}
Lingzhi Li, Jianmin Bao, Ting Zhang, Hao Yang, Dong Chen, Fang Wen, and Baining Guo.
\newblock Face x-ray for more general face forgery detection.
\newblock In \emph{Proceedings of the IEEE/CVF conference on computer vision and pattern recognition}, pages 5001--5010, 2020.

\bibitem[Li et~al.(2017)Li, Bolkart, Black, Li, and Romero]{4d}
Tianye Li, Timo Bolkart, Michael~J. Black, Hao Li, and Javier Romero.
\newblock Learning a model of facial shape and expression from 4d scans.
\newblock \emph{ACM Transactions on Graphics}, page 1–17, 2017.

\bibitem[Radford et~al.(2021)Radford, Kim, Hallacy, Ramesh, Goh, Agarwal, Sastry, Askell, Mishkin, Clark, et~al.]{radford2021learning}
Alec Radford, Jong~Wook Kim, Chris Hallacy, Aditya Ramesh, Gabriel Goh, Sandhini Agarwal, Girish Sastry, Amanda Askell, Pamela Mishkin, Jack Clark, et~al.
\newblock Learning transferable visual models from natural language supervision.
\newblock In \emph{International conference on machine learning}, pages 8748--8763. PMLR, 2021.

\bibitem[Ruiz et~al.(2023)Ruiz, Li, Jampani, Pritch, Rubinstein, and Aberman]{ruiz2023dreambooth}
Nataniel Ruiz, Yuanzhen Li, Varun Jampani, Yael Pritch, Michael Rubinstein, and Kfir Aberman.
\newblock Dreambooth: Fine tuning text-to-image diffusion models for subject-driven generation.
\newblock In \emph{Proceedings of the IEEE/CVF Conference on Computer Vision and Pattern Recognition}, pages 22500--22510, 2023.

\bibitem[Sanyal et~al.(2019)Sanyal, Bolkart, Feng, and Black]{3d}
Soubhik Sanyal, Timo Bolkart, Haiwen Feng, and Michael~J. Black.
\newblock Learning to regress 3d face shape and expression from an image without 3d supervision.
\newblock In \emph{2019 IEEE/CVF Conference on Computer Vision and Pattern Recognition (CVPR)}, 2019.

\bibitem[Song et~al.(2020)Song, Meng, and Ermon]{ddim}
Jiaming Song, Chenlin Meng, and Stefano Ermon.
\newblock Denoising diffusion implicit models.
\newblock \emph{arXiv preprint arXiv:2010.02502}, 2020.

\bibitem[Wang et~al.(2018)Wang, Wang, Zhou, Ji, Gong, Zhou, Li, and Liu]{wang2018cosface}
Hao Wang, Yitong Wang, Zheng Zhou, Xing Ji, Dihong Gong, Jingchao Zhou, Zhifeng Li, and Wei Liu.
\newblock Cosface: Large margin cosine loss for deep face recognition.
\newblock In \emph{Proceedings of the IEEE conference on computer vision and pattern recognition}, pages 5265--5274, 2018.

\bibitem[Ye et~al.(2023)Ye, Zhang, Liu, Han, and Yang]{ye2023ip-adapter}
Hu Ye, Jun Zhang, Sibo Liu, Xiao Han, and Wei Yang.
\newblock Ip-adapter: Text compatible image prompt adapter for text-to-image diffusion models.
\newblock 2023.

\bibitem[Zhang and Agrawala(2023)]{zhang2023adding}
Lvmin Zhang and Maneesh Agrawala.
\newblock Adding conditional control to text-to-image diffusion models.
\newblock \emph{arXiv preprint arXiv:2302.05543}, 2023.

\bibitem[Zhou et~al.(2022)Zhou, Chan, Li, and Loy]{zhou2022codeformer}
Shangchen Zhou, Kelvin~C.K. Chan, Chongyi Li, and Chen~Change Loy.
\newblock Towards robust blind face restoration with codebook lookup transformer.
\newblock In \emph{NeurIPS}, 2022.

\end{thebibliography}
}

\end{document}